\documentclass[final,iicol,natbib]{sn-jnl}
\usepackage{natbib}
\setcitestyle{aysep={}} 

\jyear{2022}

\begin{document}
\twocolumn[{
    \begin{center}
        \textbf{This version of the article has been accepted for publication, after peer review but is not the Version of Record and does not reflect post-acceptance improvements, or any corrections. The Version of Record is available online at: https://doi.org/10.1007/s12652-022-03849-2.}
    \end{center}
}]
\title[Article Title]{Automatic detection of cognitive impairment in elderly people using an entertainment chatbot with Natural Language Processing capabilities}

\author[1]{\fnm{Francisco} \sur{de Arriba-Pérez}}\email{farriba@gti.uvigo.es}
\equalcont{These authors contributed equally to this work.}

\author*[1]{\fnm{Silvia} \sur{García-Méndez}}\email{sgarcia@gti.uvigo.es}
\equalcont{These authors contributed equally to this work.}

\author[1]{\fnm{Francisco J.} \sur{González-Castaño}}\email{javier@gti.uvigo.es}
\equalcont{These authors contributed equally to this work.}

\author[1]{\fnm{Enrique} \sur{Costa-Montenegro}}\email{kike@gti.uvigo.es}
\equalcont{These authors contributed equally to this work.}

\affil[1]{\orgname{Information Technologies Group, atlanTTic, School of Telecommunications Engineering, University of Vigo}, \orgaddress{\street{Campus Lagoas-Marcosende}, \city{Vigo}, \postcode{36310}, \country{Spain}}}

\abstract{Previous researchers have proposed intelligent systems for therapeutic monitoring of cognitive impairments. However, most existing practical approaches for this purpose are based on manual tests. This raises issues such as excessive caretaking effort and the white-coat effect. To avoid these issues, we present an intelligent conversational system for entertaining elderly people with news of their interest that monitors cognitive impairment transparently. Automatic chatbot dialogue stages allow assessing content description skills and detecting cognitive impairment with Machine Learning algorithms. We create these dialogue flows automatically from updated news items using Natural Language Generation techniques. The system also infers the gold standard of the answers to the questions, so it can assess cognitive capabilities automatically by comparing these answers with the user responses. It employs a similarity metric with values in [0, 1], in increasing level of similarity. To evaluate the performance and usability of our approach, we have conducted field tests with a test group of 30 elderly people in the earliest stages of dementia, under the supervision of gerontologists. In the experiments, we have analysed the effect of stress and concentration in these users. Those without cognitive impairment performed up to five times better. In particular, the similarity metric varied between $0.03$, for stressed and unfocused participants, and $0.36$, for relaxed and focused users. Finally, we developed a Machine Learning algorithm based on textual analysis features for automatic cognitive impairment detection, which attained accuracy, F-measure and recall levels above 80\%. We have thus validated the automatic approach to detect cognitive impairment in elderly people based on entertainment content. The results suggest that the solution has strong potential for long-term user-friendly therapeutic monitoring of elderly people.}


\maketitle

\section{Introduction}
\label{introduction}

The United Nations has reported (World Population Prospects report\footnote{Available at {\tt https://population.un.org/wpp}, September 2021.}) that 9\% of the world population is over 65 years old, and this percentage will reach 16\% in 30 years. The population older than 80 is growing even faster, and it is expected to reach 450 million by 2050. This urges society to find innovative solutions to improve the living conditions of our elders, especially of those who live alone \citep{Callahan2014,Hancock2006}. 

A main issue of the quality of life of elderly people is the severe prevalence of cognitive impairment disorders. Affected people are mainly over 65 years old and they were 50 million in 2015, although this segment is expected to triple by 2050 \citep{Livingston2017}. Regular screening for detecting early symptoms and monitoring the progression of these disorders have been considered beneficial for treatment planning and patient autonomy \citep{Borson}. However, it has also been observed that cognitive impairment assessment in primary care systems is inefficient \citep{Lopponen2003, Boise2004}.

As discussed in Section {\ref{related_work}}, even though existing telecare set-top-boxes and gateways have increasingly intelligent capabilities, they still do not communicate autonomously with the elderly using natural language. This is also the case of the solutions for cognitive evaluation, which mostly rely on sets of predefined manual tests.

Given the industrial gap, and as demonstrated by the analysis of the state of the art in Section {\ref{related_work}}, we propose a novel conversational system for entertainment and therapeutic monitoring of elderly people that relies on {\sc nlp} techniques and Machine Learning for empathetic chatbot behaviour generation and user-transparent automatic assessment.

From the perspective of the target users, the elderly, the main priority of any information system should be alleviating loneliness, whether the system has embedded cognitive monitoring capabilities or not. Accordingly, we want our solution to be perceived as a friendly intelligent assistant to access Internet media, that is, a conversational system that reads news. These will be interspersed with brief dialogues to subtly guide the users through a series of questions to gather their interests and evaluate their understanding of the information they have just consumed, which includes word category understanding and short-term memory, to evaluate cognitive impairment \citep{Loewenstein2004, Crocco2014}. 

Conversational systems seem an adequate approach for this purpose. These software programs allow for human interaction with a machine using written or spoken natural language \citep{Shawar}. Ideally, the dialogue should be empathetic \citep{Fung, Rashkin}, a still distant goal even after the recent advances in Artificial Intelligence ({\sc ai}) and in Natural Language Generation ({\sc nlg}) techniques. Before virtual companions \citep{Shum2018} become a reality, entertainment through interesting information will be more feasible. In this vein, it is well known that elders feel accompanied by simply listening to news in their background \citep{Ostlund2010}.

More in detail, our system reads recent news items and generates questions about them. At the same time, in order to automatically evaluate cognitive capabilities, the system measures the similarity between user answers and a gold standard \citep{Yang2005,corley2005measuring,li2006,feng2008sentence} that is automatically generated from the news. To validate our approach, we have defined an answer similarity metric and we have performed tests on a sample of 30 patients of Asociaci\'on de Familiares de enfermos de Alzheimer y otras demencias de Galicia ({\sc afaga}\footnote{Available at {\tt https://afaga.com}, September 2021.}, the Galician Association of Relatives of Patients with Alzheimer’s and other Dementias). This public association seeks to improve the quality of life of Alzheimer patients, provides guidance and information to relatives and to the public, and makes society aware of this reality to achieve a broader and more effective response. It collaborates actively in research on cognitive impairments.

The rest of this paper is organised as follows. Section \ref{related_work} reviews related work and lists our contributions. Section \ref{system_architecture} describes our conversational system for entertainment and user-transparent cognitive assessment. Section \ref{evaluation} presents our case study and the results of our word similarity approach to automatically evaluate cognitive impairment. Finally, Section \ref{conclusions} concludes the paper.

\section{Related Work}
\label{related_work}

The design of intelligent systems \citep{Wilamowski2015} is a relevant research field, including for example industrial \citep{Irani2014,Ngai2014}, military \citep{MaSum2013,Yoo2014} and social \citep{Magnisalis2011,Bernardini2014,Adamson2019} applications. In the telecare domain, personal assistants \citep{Matsuyama2016,Lopez2018} and artificial companionship \citep{Chumkamon2016,Abdollahi2017} are relevant.

Regarding the automatic detection of health conditions, there exists a wealth of work based on Machine Learning, such as \cite{Ghoneim2018}, on a smart healthcare framework to detect medical data tampering; \cite{Sedik2021}, on the analysis of the outbreak of {\sc{covid}}-19 disease; \cite{Ahmed2021}, on an unsupervised Machine Learning approach to predict data-types attributes for optimal processing of telemedicine data, including text and images; and \cite{Sarrab2021}, on the real-time detection of abnormalities in streamed data from IoT sensors (furthermore, in \cite{Masud2021}, a mutual authentication and secret key establishment protocol was proposed to protect medical IoT networks). Our research contributes to automatic smart telecare solutions based on Machine Learning.

However, despite the huge advances in {\sc ai} for artificial reasoning and problem-solving, interactions are still far from human-like. Most recent {\sc ai} systems have partial understanding of natural language and lack cognitive capabilities to enrich the communication with context-dependent information \citep{Skjuve2019HelpInteraction}. Specifically, existing solutions for intelligent conversational systems are based on retrieval-based methods \citep{Yasuda2014,Wu2018}, which select the best candidate among a predefined set of alternative responses, and generation-based procedures \citep{Oh2017,Su2018,Baby2018}, which rely on {\sc nlp} techniques to create human-like written or oral dialogue flows. Note that natural language is especially beneficial for user interfaces for its spontaneity and friendliness \citep{Liu2018}.

Among the existing intelligent conversational systems, we can mention Google Duplex \citep{Lindgren2011} and the Neural Responding Machine in \cite{Shang2015} based on Recurrent Neural Networks. Moreover, in \cite{Wen2017} the authors presented a dialogue system based on the pipelined Wizard-of-Oz framework, which, unlike other approaches in the literature, can make assumptions. Regarding linguistic knowledge, in \cite{Wang2015} and \cite{Wu2018} syntactic features were used to generate coherent and human-like texts. Newscasters \citep{Matsumoto2007JournalistWorld}, which, as previously said, generate a feeling of companionship \citep{Ostlund2010}, do not sustain dialogues with end users. Conversational systems have already been considered for entertainment and healthcare \citep{Noh2017,Su2018}, although still at an early stage.

Regarding existing conversational systems for entertainment \citep{Johnson2016,Correia2016,Aaltonen2017}, we highlight EduRobot \citep{Cahyani2017} which can sing and tell stories, although it is not oriented to the specific needs of the elderly. It has been suggested to make conversational systems more appealing by modelling their interfaces as pets or avatars \citep{Sharkey2012}. Unlike EduRobot, RobAlz \citep{Salichs2016} has been specifically devised for this audience, but it has no therapeutic diagnostic capabilities. 
Few existing intelligent systems for senior healthcare \citep{Foroughi2008,Hsu2009,Suryadevara2012,Tseng2013,Samanta2014,Wang2016} have human-computer communication capabilities. The system by \cite{Yasuda2014} for people with dementia is an exception, but its communication capabilities are very limited: it just selects questions and answers among 120 pre-set options. Therefore, even though there still is incipient work on digital tools for therapeutic monitoring of people with dementia and other impairments, manual tests (written and task-based neurological and neuropsychological assessments with caretaker supervision) are typically used. For example, the Mini-Mental State Examination \citep{Ridha2005} is a cognitive test on orientation, immediate memory, calculation, attention and comprehension, to cite some tasks, which produces scores about dementia levels. The Mini Cognitive Assessment Instrument \citep{Milne2008} includes a verbal memory task and a clock drawing test. Finally, the {\sc camdex} test \citep{Ball2004} is another standardised manual tool for the diagnosis of mental disorders, which is especially suitable for the early detection of dementia. It asks questions related to memory, personality, general mental and intellectual functioning, and judgement. It also considers specific symptoms and the medical histories of the users and their families. Note that the white-coat effect is a major concern in all these manual approaches, apart from the fact that they are time-consuming and require professional expertise. 

Previous research has proved the benefits of combining dichotomous questions (also named closed questions) with essay questions\footnote{Dichotomous or closed questions have binary yes/no answers. Essay questions allow users to express themselves freely.} to mitigate the white-coat effect \citep{Ridha2005,Echeburua2017}. We remark the interest of inserting distracting questions before attention-demanding questions in cognitive tests \citep{Ridha2005}. Altogether, this combination allows evaluating cognitive impairment less intrusively \citep{Ball2004,Ridha2005,Milne2008}. To the best of our knowledge, our proposal is the first intelligent system that embeds user-transparent, automatic cognitive assessment into a newscaster system that sustains dialogues with elderly people, based on {\sc nlp} techniques for chatbot behaviour generation and human-machine communication.

We close this section with a review of related industrial initiatives, which further backs the social relevance of the field under study.

For instance, the Carelife system by Telev\'es\footnote{Available at {\tt http://televescorporation.com/areas-de-
negocio/sociosanitario}, September 2021.} analyses personal routines from sensor data for custom home care. Its home gateway can be extended via peripherals, such as biomedical devices, as well as via software. Doro launched SmartCare\footnote{Available at {\tt https://www.doro.com/es-es/care}, September 2021.} in 2018. It includes a home gateway and home sensors to detect behavioural changes. However, neither these systems nor the {\sc sam} Robotic Concierge by Luvozo\footnote{Available at {\tt https://luvozo.com}, September 2021.} have built-in intelligence to communicate with the elderly using natural language.

Regarding general purpose domestic robots, there are examples such as ZenBo by Asus\footnote{Available at {\tt https://zenbo.asus.com}, September 2021.}, with video surveillance, Internet shopping and agenda features. Its interactions are rather rigid. It can understand vocal orders, but it has no empathetic capabilities, nor is it tailored to the needs of the elderly.

We believe that a feasible path towards a next generation of intelligent conversational telecare systems is the augmentation through software of current platforms such as Carelife and Smartcare by relying on their simple voice interfaces (which, nowadays, caregivers employ to call the users), without any additional hardware add-ons. Some platforms are already open in this aspect. For example, Buddy, by European Blue Frog Robotics\footnote{Available at {\tt https://buddytherobot.com/en/buddy-the-
emotional-robot}, September 2021.}, allows third developers to create new applications and distribute them via its store. We are neither aware of any application of this robot to entertain and monitor elderly people, nor of any intelligent functionalities.

Regarding solutions focusing on cognitive evaluation, we must mention three approaches based on manual tests (written and task-oriented neurological and neuropsychological assessments), none of which employ automatic {\sc nlp} techniques. Neurotrack\footnote{Available at {\tt https://neurotrack.com}, September 2021.} is a set of cognitive tests to evaluate, monitor and strengthen brain health to reduce the risk of Alzheimer and other dementias. Mezurio\footnote{Available at {\tt https://mezur.io}, September 2021.} provides support to interactive data acquisition as a baseline for detecting individuals at risk of developing Alzheimer's disease. Altoida\footnote{Available at {\tt http://altoida.com}, September 2021.} tests the functional and cognitive skills of patients with a Machine Learning algorithm but, as the previous two solutions, it is entirely based on a set of predefined tests and it does not have any bidirectional communication capabilities in natural language.

\section{System Architecture}
\label{system_architecture} 

\begin{figure*}[ht!]
\centering
\includegraphics[width=\textwidth]{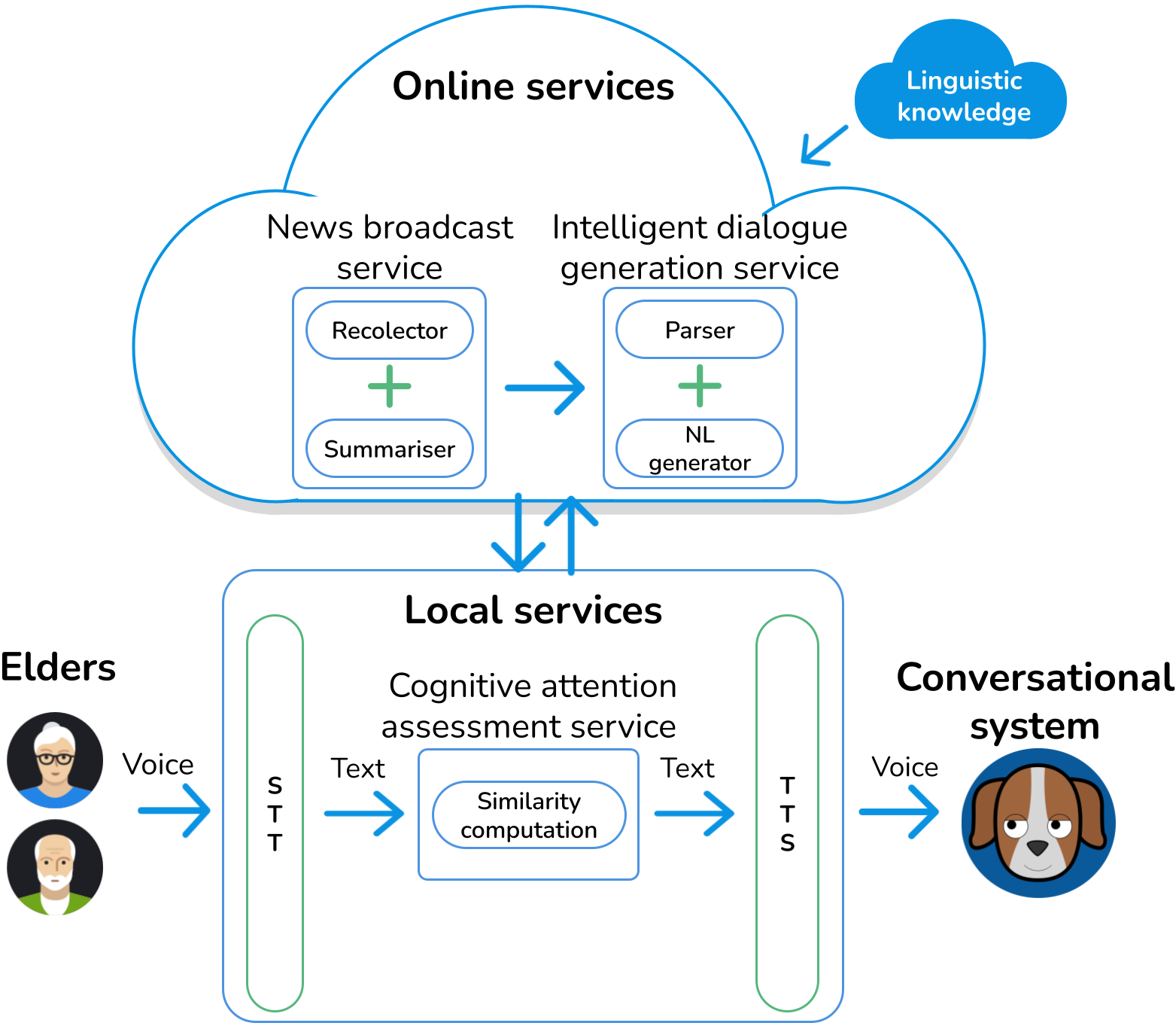}
\caption{\label{architecture} System architecture.}
\end{figure*}

We present a novel intelligent system specially designed for therapeutic monitoring of elderly people with different levels of cognitive impairments or on the verge of suffering them. The users perceive the system as a news broadcasting service, with which they interact by voice from time to time. Therapeutic monitoring is embedded as a user-transparent functionality. Figure \ref{architecture} illustrates the main modules of the system, on which we will elaborate in the next sections. They include online (news broadcast service and intelligent dialogue generation system) and local services (Android\footnote{Available at {\tt https://www.android.com}, November 2021.} application with cognitive attention assessment service). Online services modules were implemented using the Eclipse\footnote{Available at {\tt https://www.eclipse.org}, November 2021.} Integrated Development Environment ({\sc ide}) tool and Java 1.8 programming language\footnote{Available at {\tt https://www.java.com}, November 2021.}. They were deployed on a Tomcat\footnote{Available at {\tt http://tomcat.apache.org}, November 2021.} server to be made available through a \textsc{rest api}, which was programmed using the Jersey library\footnote{Available at {\tt https://eclipse-ee4j.github.io/jersey}, November 2021.}. The Android application (for Lollipop operating system or higher regarding devices compatibility) was developed with Android Studio\footnote{Available at {\tt https://developer.android.com/studio}, November 2021.}.

The system transforms Spanish speech into text and vice versa as input and output data ({\sc stt}/speech-to-text and {\sc tts}/text-to-speech boxes in Figure \ref{architecture}). For this purpose, it employs the Google Voice Android Software Development Kit ({\sc sdk}) library\footnote{Available at {\tt https://developer.android.com/reference/
android/speech/Speech
Recognizer}, September 2021.}. 

Regarding system activation, we use voice commands and facial recognition. For implementing the latter we employed the OpenCV library\footnote{Available at {\tt https://opencv.org}, September 2021.} and an eye-sensing train data set\footnote{Available at {\tt https://github.com/opencv/opencv/blob/
master/data/lbpcascades/lbpcascade\_frontalface.xml}, September 2021.}
(note that previous works have also exploited sophisticated schemes for this purpose, apart from voice commands \citep{Alsmirat2019}).

\begin{figure*}[ht!]
\centering
\includegraphics[scale=0.085]{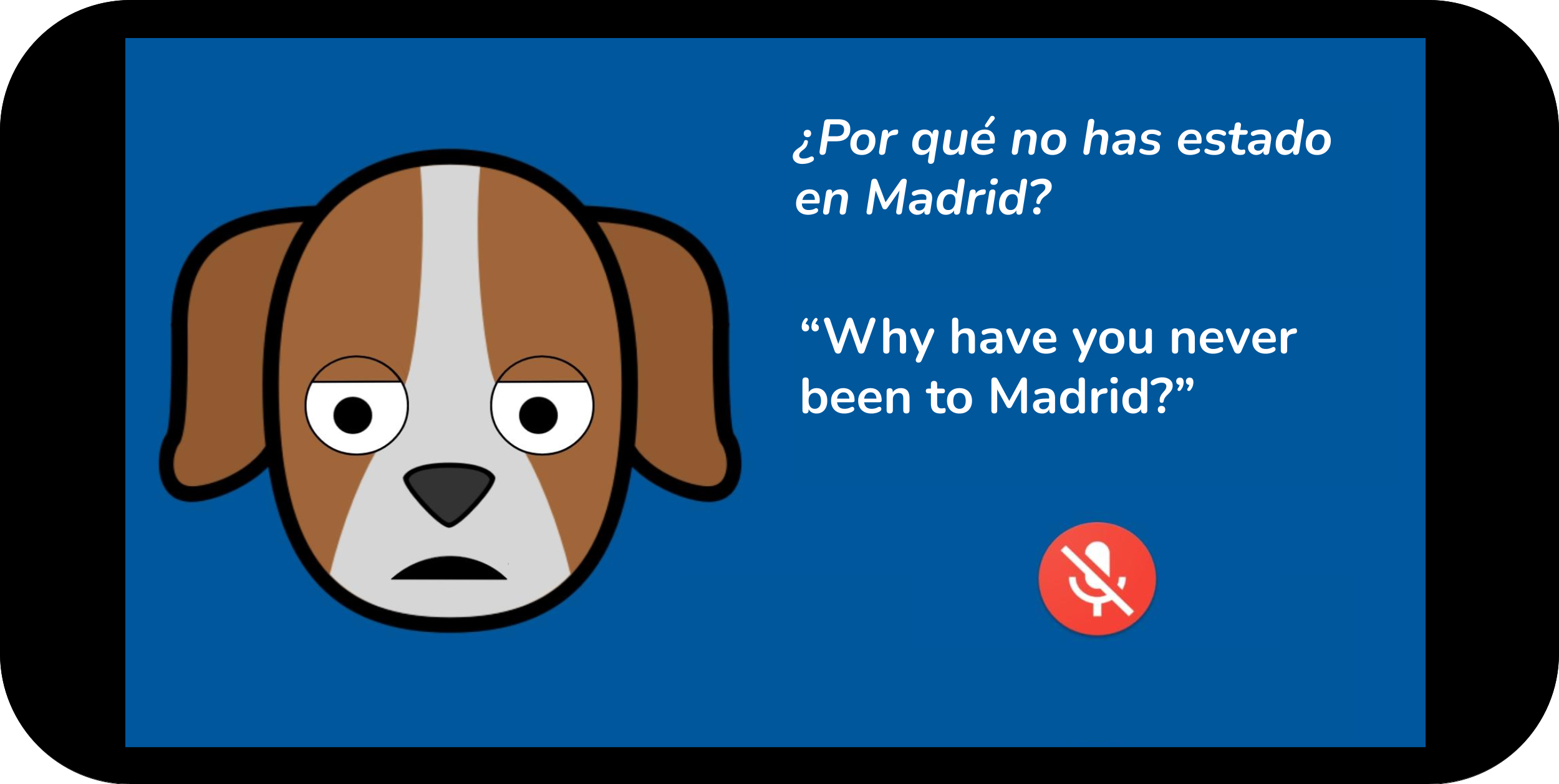}
\caption{\label{chatbot_interface} User interface.}
\end{figure*}

As in \cite{Wang2020}, we decided to combine text and images, in our case to help the users focus during the dialogue stages. Specifically, the dialogue with the users is guided with basic graphic indicators (see Figure \ref{chatbot_interface}): the screen displays a ``muted'' or ``open'' microphone to indicate the system's or user's turn to speak. Moreover, the ``facial'' expressions of the avatar of the conversational system provide empathetic feedback to the users based on text sentiment analysis. Finally, note that the user interface is an animated dog, simulating a pet as suggested in literature \citep{Sharkey2012}.

To ensure short response times, instead of querying external systems, the system relies on a MongoDB database\footnote{Available at {\tt https://www.mongodb.com}, September 2021.} in our own server containing all the necessary linguistic knowledge. 

\subsection{News Broadcast Service}
\label{news_broadcast_service}

The news broadcast service requires a varied and updated set of news to engage the target users. This news is periodically extracted from the Application Programming Interface ({\sc api}) of the Spanish National Radio and Television (RTVE\footnote{Available at {\tt https://www.rtve.es/api}, September 2021.}) channel with a {\sc get} query, using the {\tt tematicas} and {\tt noticias} {\sc api} services, to gather news items on specific topics. This task is performed in the background owing to the requirements of the {\sc{api}}, and it hides news processing to the target users by generating pre-saved news items on a daily basis. The content retrieved from the {\sc{api}} is saved into a MongoDB structure using a {\sc{json}} file. Date and topic features are exploited for indexing and searching. As a result, the News Broadcast Service delivers content immediately from the user's point of view.

News are arranged into five categories: economy, politics, science, society and sports, all of them at national level except for politics and society, which focus on Galicia (Spain) in our case, since proximity is appealing to elderly people. 

Table \ref{news} presents an example of a social news item. We provide the user with a summary of each piece of news by extracting the lead paragraph (for that purpose, we first split the news items into paragraphs and take the first paragraph after the title).

\begin{table*}[ht!]
\centering
\caption{\label{news} Example of news content.}
\begin{tabular}{cc}
\hline
\bf Topic & \bf News item\\\hline
Society & \begin{tabular}[c]{@{}p{10cm}@{}}
{\it La Polic\'ia Nacional ha desarticulado una peligrosa red de narcotraficantes que operaba en Galicia, Madrid y Alicante en una operaci\' on en la que han sido detenidas diez personas con numerosos antecedentes por delitos graves y se han intervenido 1.000 kilos de coca\'ina, 500.000 euros y diez veh\'iculos de lujo.}\\`National Police has dismantled a dangerous drug trafficking ring that operated in Galicia, Madrid, and Alicante in an operation in which ten people with a wide criminal record have been arrested, and 1,000 kilos of cocaine, 500,000 euros and ten luxury cars have been seized.'
\end{tabular}\\\hline
\end{tabular}
\end{table*}

\subsection{Automatic Question Generation}
\label{dialogue}

Our question generation system combines linguistic knowledge from our aLexiS lexicon \citep{Garcia-Mendez2018,Garcia-Mendez2019} with the Name Entity Classification ({\sc nec}) functionality from Freeling \citep{Atserias06,PadroEtAl12}. The former is saved in a MongoDB database to reduce the response time of the chatbot. Besides, the {\sc nec} process is executed in the background because of the complexity of the linguistic analysis of the news.

These two resources allow extracting and identifying personal names, organisations and locations, which together constitute valuable data for question generation. More specifically, thanks to the linguistic information in aLexiS, our conversational system can adjust features such as the gender, number, person, and tense of the questions it generates.

As previously said, we follow the strategy of combining dichotomous and essay questions to reduce the white-coat effect and create a more relaxed atmosphere.

\begin{itemize}
 \item To generate dichotomous questions, we rely on the {\sc nec} functionality of Freeling to extract personal names and locations from the news. This produces questions such as those in Table {\ref{tab:dichotomous_questions}}. The system always generates four similar options for each question, and one of them is picked at random and presented to the user. Then, depending on the user's answer, the system poses the next question as indicated in Table {\ref{tab:dichotomous_questions_next}}.

\begin{table*}[ht!]
\centering
\caption{\label{tab:dichotomous_questions} Example of dichotomous questions by different {\sc nec} results.}
\begin{tabular}{ll}\toprule
\bf NEC & \bf Example of dichotomous questions\\\hline
\multirow{2}{*}{People} & {\it ¿Has o\'ido hablar de ENTIDAD?} \\
& `Have you ever heard about ENTITY?'\\
\multirow{2}{*}{Location} & {\it ¿Has estado alguna vez en ENTIDAD?}\\
& `Have you ever been to ENTITY?'\\\bottomrule 
\end{tabular}
\end{table*}

\begin{table*}[ht!]
\centering
\caption{\label{tab:dichotomous_questions_next} Example of dichotomous questions by different {\sc nec} results depending on the user's response.}
\begin{tabular}{lll}\toprule
\bf Answer & \bf NEC & \bf Example of dichotomous questions \\\hline
\multirow{4}{*}{Yes} & \multirow{2}{*}{People} & {\it ¿Qu\'e datos conoces de la vida de ENTIDAD?} \\
& & `What facts do you know about ENTITY's life?'\\
& \multirow{2}{*}{Location} & {\it Cu\'entame qu\'e es lo que m\'as te ha gustado de ENTIDAD}\\
& & `Tell me what you liked the most about ENTITY'\\

\multirow{4}{*}{No} & \multirow{2}{*}{People} &{\it ¿Por qu\'e ha saltado a los medios de comunicaci\'on ENTIDAD?}\\
& & `Why has ENTITY jumped into the media?'\\
& \multirow{2}{*}{Location} & {\it ¿Por qu\'e no has estado en ENTIDAD?}\\
& & `Why have you never been to ENTITY?'\\

\multirow{4}{*}{{\sc n/a}} & \multirow{2}{*}{People} & {\it ¿Qu\'e visi\'on de ENTIDAD nos dan los medios de comunicaci\'on?}\\
& & `Which is the view of ENTITY in the media?'\\
& \multirow{2}{*}{Location} & {\it ¿Podr\'ias contarme algo relacionado con ENTIDAD?}\\
& & `Could you tell me anything about ENTITY?' \\\bottomrule
\end{tabular}
\end{table*}
\end{itemize}

Regarding the extraction of the gold standard answers for the aforementioned types of questions, we obtain these data with Freeling syntactic parsing. Take the sentence `National Police has dismantled a dangerous drug trafficking ring that operated in Galicia, Madrid, and Alicante' as an example. The corresponding essay question using our system is `Who has dismantled a dangerous drug trafficking ring?', and the correct answer it produces as a reference is the noun phrase that precedes the verb, `National Police'. Note that the best answer is obtained by extracting the noun phrases that precede the verb for `who' questions, as in our example. On the other hand, for `what' questions, we use the noun phrase that precedes the verb plus the verb itself. Take the sentence `The Government will automatically extend the social electric bonds until September 15th' and its associated question `What does the news say on September 15th will happen?' as an example of `what' question handling. The correct produced answer is `It will automatically extend social electric bonds'. Finally, for the question `Which places does the news item mention?', the correct answer is produced by extracting all location entities from the news content using the {\sc nec} functionality by Freeling. Note that in the first example about the drug trafficking ring, the correct answer is {\it Galicia}, {\it Madrid} and {\it Alicante}.

Both the generation of questions and the extraction of gold standard answers are performed in the background, after the daily news-gathering process (see Section {\ref{news_broadcast_service}}), using the same indexing scheme.

By combining dichotomous and essay questions, the conversational system establishes a dialogue with the end users. Each dialogue is composed of the following three stages:

\begin{itemize}
 \item News: prior to the dialogue, the conversational system presents the news item.
 
 \item User-centred questions: a dichotomous question followed by two essay questions to distract the user.
 
 \item Attention-demanding question: a last essay question on key aspects of the news item. It allows assessing if the user understood the news piece, if he/she was focused during the conversation, and his/her short-term memory.
\end{itemize}

\begin{figure*}[ht!]
\centering
\includegraphics[scale=0.23]{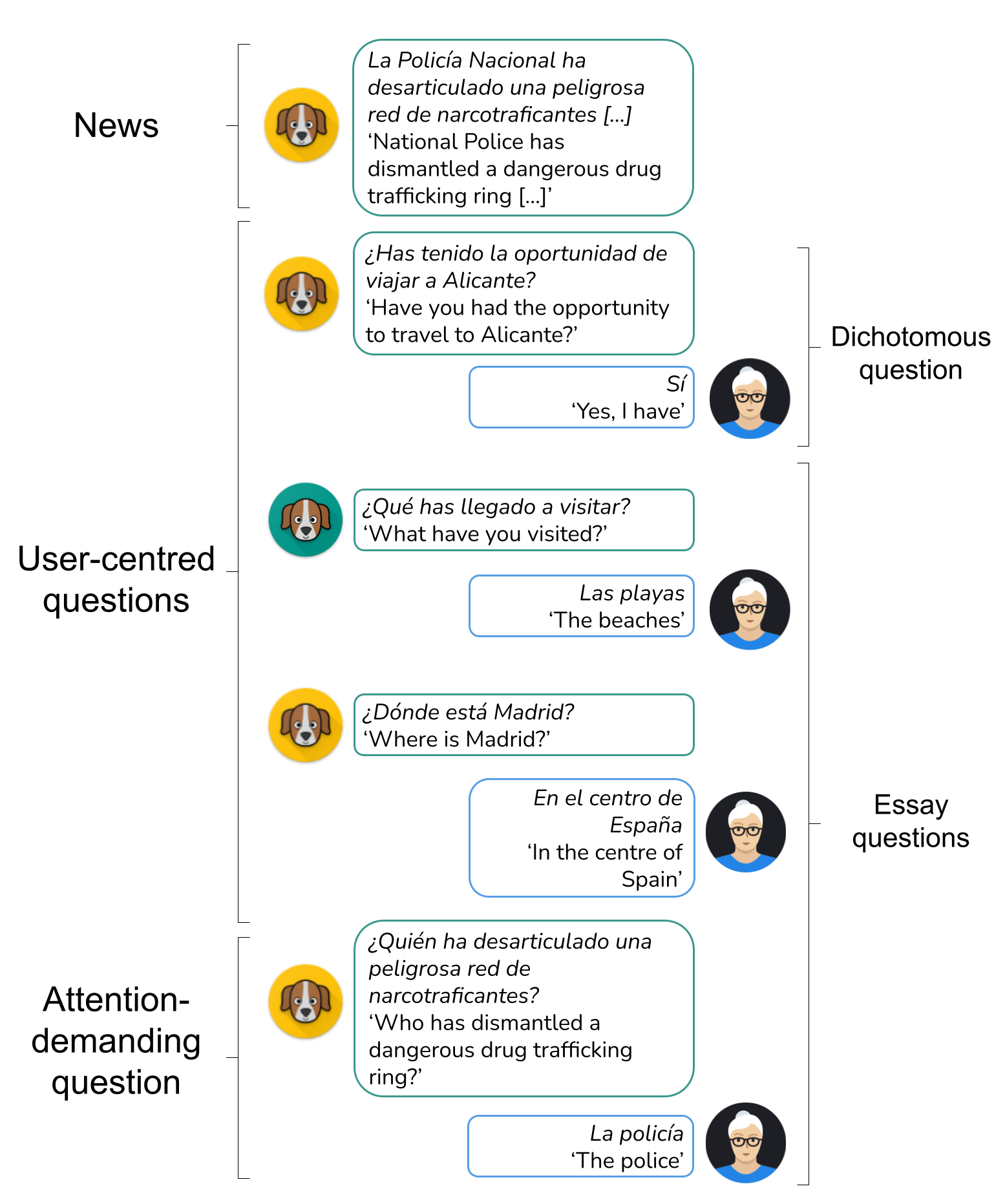}
\caption{\label{flow_diagram} Real conversation example.}
\end{figure*}

Figure \ref{flow_diagram} shows an example of a real dialogue according to this structure. To keep the user engaged, most questions are related to the news item (avatar marked in yellow).

The resulting user's utterances are the input to the cognitive attention assessment service, which calculates the accuracy of the user answers by comparing them with the gold standard responses. We describe it in the next section.

\subsection{Cognitive Attention Assessment Service}
\label{attention}

We employ the lexical Multilingual Central Repository ({\sc mcr}\footnote{Available at {\tt http://adimen.si.ehu.es/web/MCR}, September 2021.}) database \citep{GonzalezEtAl12}, which integrates the Spanish WordNet into the EuroWordNet framework, to obtain the semantic classification of the adjectives, adverbs, nouns and verbs in the news piece. For that purpose, we extract from {\sc mcr} three semantic categories corresponding to Adimen {\sc sumo}, WordNet Domains and Top Ontology hierarchies for nouns and verbs, and Top Ontology hierarchies for adjectives and adverbs, since there is less information available for these two lexical categories \citep{pedersen2004wordnet}. From {\sc mcr} we also gather holonyms, hypernyms, hyponyms, meronyms, synonyms and related data for nouns and verbs, and only synonyms for adjectives and adverbs. Table \ref{semantics} shows an example for noun {\it monta\~na} `mountain'.

This linguistic knowledge was added to the aLexiS lexicon within the same {\sc{json}} indexing scheme, which did not affect response time performance. 

\begin{table*}[ht!]
\centering
\small
\caption{\label{semantics} Semantic data from {\sc mcr} for noun {\it monta\~na} `mountain'.}
\begin{tabular}{ll}
\hline
{\bf Feature} & {\bf Value} \\\hline
Word & {\it monta\~na} `mountain' \\
WordNet Domain & object \\ 
Adimen Sumo & LandArea \\ 
Top Ontology & geography, geology \\
Holonym & - \\ 
Hypernym & {\it elevaci\'on} `elevation' \\ 
Hyponym & {\it Alpes} `Alps', {\it volc\'an} `volcano' \\ 
Meronym & {\it vertiente} `slope', {\it cumbre} `summit'\\ Synonym & {\it monte} `hill' \\
Related & {\it monta\~nero} `mountaineer', {\it alpinista} `alpinist',\\
& {\it monta\~noso} `mountainous', {\it montuoso} `mountainous' \\ \hline
\end{tabular}
\end{table*}

We obtain the final score for each response as a weighted average, as follows:

\begin{align}
\label{eq:0}
 sim=0.8 \frac{\sum_{i=1}^{N_{noun}} {noun^*_i} + \sum_{i=1}^{N_{verb}} {verb^*_i}}{N_{noun} + N_{verb}} \notag\\ + 0.2 \frac{\sum_{i=1}^{N_{adj}} {adj^*_i} + \sum_{i=1}^{N_{adv}} {adv^*_i}}{N_{adj} + N_{adv}} 
\end{align}

Where:

\begin{itemize}
 \item $N_x$ represents the number of words of lexical category $x$ in the ideal response.
 \item For the $i$-th word of category $x$ in the ideal answer, we calculate its similarity with all words within the same lexical category $x$ in the user's response, and we take the highest value as $x^*_{i}$.
\end{itemize}

Note that nouns and verbs are considered more relevant than adjectives and adverbs in expression \eqref{eq:0}. This choice is supported by the fact that the former generally carry most semantic information in a sentence, whereas the latter provide nuances \citep{feng2008sentence, corley2005measuring}.

To obtain $x^*_i$ we adapted the method by Yang et al. \citep{Yang2005} to calculate the similarity $s(word_1, word_2)$ between words $word_1$ and $word_2$:

\begin{align}\label{eq:1}
 s{(word_1,word_2)} = (1-\gamma) \alpha_s \beta^{d(word_1,word_2)} + \gamma
\end{align}

Where:

\begin{itemize}
 \item $\alpha_s = 0.9$ if the words are synonyms and $0.85$ otherwise. 
 \item $d(word_1,word_2) = 0$ if the words are holonyms, hypernyms, hyponyms, meronyms or synonyms, related to the same hierarchy category or belonging to it. Otherwise, $d(word_1,word_2)$ is WordNet's shortest path between $word_1$ and $word_2$.
 \item $\beta$ acts as a depth factor that decreases similarity exponentially, to the power of the number of hierarchical steps separating the two concepts. In the tests in Section \ref{evaluation} we set it to $0.7$.
 \item $\gamma$ is explained below.
\end{itemize}

Note that certain fairly similar words in the same WordNet domain category will have a very low similarity value with our method (less than $0.4$). Consider for example the pair ({\it madera} `wood', {\it cart\'on} `cardboard'), whose similarity is $0.27$, a low value taking into account that both terms define materials and share WordNet domain category `substance'; and pair ({\it panadero} `baker', {\it maestro} `teacher'), with a similarity of $0.10$, although these words represent professions and belong to the same WordNet domain category `person'.

To avoid this issue, we define correction factor $\gamma$ in expression \eqref{eq:1}. By default $\gamma=0$, except for the following two cases:

\begin{itemize}

\item For word pairs that belong to the same WordNet domain category, $\gamma=0.25$. After applying this to the two previous examples, their similarities grow to $0.45$ and $0.33$ (from $0.27$ and $0.10$), respectively.

\item For all terms with the same stem that have not been already classified as synonyms, with a similarity of $0.85$ or less, $\gamma=0.5$. This is because they belong to the same word family, and their similarity should be even higher for our purposes than in the previous case. For instance, the pair ({\it flor} `flower', {\it florista} `florist') would have a similarity value of $0.15$ for $\gamma=0$, but it becomes $0.58$ after applying $\gamma=0.5$.

\end{itemize}

The goal of these corrections is improving coherence, by defining a ground truth for word similarity \citep{li2006}. This was tuned by considering the value range ($0.3$, $0.6$) for similarity scores in cases such as {\it aut\'ografo} `autograph' versus {\it firma} `signature' and {\it coj\'in} `cushion' versus {\it almohada} `pillow'.

Moreover, we pay special attention to the treatment of numbers. Given the fact that our goal is to assess the understanding of the news, we must take into account that even people with healthy minds seldom retain the exact quantities they have just heard. For example, after listening to the sentence `there were 2569 casualties', a person will likely remember `there were over 2500 casualties'. For this reason, we generate all the possible numbers a given quantity can be rounded to by dividing it by powers of ten, and we assign a $0.7$ similarity score if any of them produces a match. For instance, if the correct amount is 2569, the possible right answers that would be assigned a $0.7$ similarity score are 2000, 2500, 2560, 2570, 2600 and 3000. On top of that, if the words `over' or `under' are chosen correctly, a similarity score of $0.9$ is assigned. Going back to the previous example, if the user's reply is `there were over 2500 casualties', it receives a $0.9$ similarity score, whereas if it is `there were 2500 casualties' it gets a $0.7$ similarity score, since it reflects less understanding of the original information.

As a final example, take the sentence {\it un profesor llev\'o papel en blanco a su hogar en la monta\~na} `a teacher took blank paper to his home in the mountain' as the ideal response to a question. The following three user answers produce different results:
\begin{itemize}
 \item {\it Un hombre llev\'o folios blancos a su casa del monte} `a man took blank paper sheets to his house on the hill': this reply uses different words than the original, but keeps the same meaning. It would obtain a similarity score of $0.8$.
 \item {\it Un hombre sac\'o madera blanca de su apartamento} `a man took white wood from his apartment': this sentence has lost most original meaning, although some common concepts remain. It would obtain a similarity score of $0.51$. 
 \item {\it Un hombre rompi\'o una silla en una tienda} `a man broke a chair in a shop': this sentence has none of the original meaning left. It would obtain a similarity score of $0.25$ (note that the subject was correctly inferred).
\end{itemize}

The $sim$ metric allows the automatic evaluation of comprehension skills from the questions and their corresponding gold standard answers, and the procedure reflects the level of concentration during the conversation and the reliability of short-term memory.

\section{Experimental Results and Discussion}
\label{evaluation}

In this section, we present the validation tests to assess the effectiveness of our approach to determine cognitive impairment levels.

\subsection{Hardware}

We ran the system on a server with the following characteristics:

\begin{itemize}
 \item Operating System: Ubuntu 18.04 LTS 64 bits
 \item Processor: Intel Xeon CPU E5-2620 v2 2.1 GHz
 \item RAM: 64GB DDR3
 \item Disk: 3 Tb
\end{itemize} 

\subsection{Case Study}

The case study comprises two experimental scenarios. The first experiment (Section {\ref{sim_study}}) studies the {$sim$} metric presented in Section {\ref{attention}} as a tool to assess abstraction capabilities, for different user profiles and levels of cognitive impairment. The second experiment (Section {\ref{auto_impairment}}) evaluates Machine Learning algorithms to automatically detect cognitive impairment in the users under study.

These experiments were divided in ``sessions'', a session being a particular newscast and its associated dialogue with an elderly person. The profiles of the participants in these sessions were characterised as follows:

\begin{itemize}
 \item At least 60 years old.
 \item Technological skills and hearing problems: existing or not.
 \item Study levels: basic (high education or less) or superior (bachelor's, master's and doctoral degrees).
 \item Cognitive impairment level: absent, mild or severe, as established for the case study by gerontology experts from {\sc{afaga}}.
 \item Stress: yes or no.
 \item Focus: yes or no.
\end{itemize}

The experiments lasted for three months and involved 30 users 75.73 $\pm$ 6.60 years old (average $\pm$ standard deviation). All the users involved in the experiments are patients in the occupational therapy workshops of {\sc afaga}. The tests were conducted under the supervision of their caregivers. For annotating cognitive impairment, {\sc afaga} applied the Spanish version \citep{DiazMardomingo2008} of the Global Deterioration Scale standard methodology \citep{Reisberg1982}. Specifically, 57\% of the participants had cognitive impairment to some extent (40\% mild, 17\% severe) and the rest were mentally healthy. We used this methodology as manual baseline for the comparison with our automatic Machine Learning detection approach in Section {\ref{auto_impairment}}.

In each individual experiment, we registered the characteristics of the user. Tables \ref{usersdata} and \ref{newsdata} shows the session registration sheet, which was filled by the caregivers, and a real example of a session, with its newscast content and its associated interview.

\begin{table*}[ht!]
\centering
\small
\caption{\label{usersdata} Session registration sheet.}
\begin{tabular}{|l|}
\hline
\begin{tabular}[c]{@{}p{10cm}}{\bf Personal information}\\ \\ 
\hspace{0.5cm} User ID: \\ 
\hspace{0.5cm} Age: \\ \\

\hspace{0.5cm} Are you familiar with the use of new technologies?\\
\hspace{2cm} $\bigcirc$ Yes \hspace{2cm} $\bigcirc$ No \\ \\

\hspace{0.5cm} Do you suffer from hearing problems?\\
\hspace{2cm} $\bigcirc$ Yes \hspace{2cm} $\bigcirc$ No \\ \\

\hspace{0.5cm} Level of education\\
\hspace{2cm} $\bigcirc$ Basic \hspace{1.75cm} $\bigcirc$ Superior \\ \\

{\bf Session data} (filled by the caregiver)\\ \\ 

\hspace{0.5cm} What is the level of cognitive impairment?\\
\hspace{2cm} $\bigcirc$ Absent \hspace{1.55cm} $\bigcirc$ Mild \hspace{1.4cm} $\bigcirc$ Severe\\ \\

\hspace{0.5cm} Frame of mind\\
\hspace{2cm} $\bigcirc$ Happy \hspace{1.6cm} $\bigcirc$ Normal \hspace{1.1cm} $\bigcirc$ Sad\\ \\

\hspace{0.5cm} Was the user stressed?\\
\hspace{2cm} $\bigcirc$ Yes \hspace{2cm} $\bigcirc$ No \\ \\

\hspace{0.5cm} Was the user focused?\\
\hspace{2cm} $\bigcirc$ Yes \hspace{2cm} $\bigcirc$ No \\ \\

{\bf Additional information}\\ \\ \\

\end{tabular} \\ \hline
\end{tabular}
\end{table*}

\begin{table*}[ht!]
\centering
\small
\caption{\label{newsdata} News item for session 1.}
\begin{tabular}{|l|}
\hline
\begin{tabular}[c]{@{}p{11cm}}{\bf News 1: politics}\\ \\ 
User ID: \\ \\
`Teresa Ribera has announced this Thursday that the Government will automatically extend electricity social bonds until September 15th to maintain the protection of people in a condition of energy vulnerability and to guarantee basic supplies at home due to the coronavirus crisis.'\\ \\
{\bf Question 1}: Does the name Teresa Ribera sound familiar to you?\\ \\
\hspace{1cm} Answer:\\ \\
\hspace{2cm} Yes $\rightarrow$ {\bf Question 1.1}: What facts do you know about the life of Teresa Ribera?\\ \\ 
\hspace{2cm} No $\rightarrow$ {\bf Question 1.2}: What do you associate the name of Teresa Ribera to?\\ \\
\hspace{2cm} Not available $\rightarrow$ {\bf Question 1.3}: Why do you think people like the name Teresa?\\ \\
{\bf Question 2}: What is the meaning of the word `Government'?\\ \\ 
{\bf Question 3}: What happened on September 15th according to the news?\\ \\ 
{\bf Question 4}: Do you consider this news item interesting?\end{tabular} \\ \hline
\end{tabular}
\end{table*}

In detail, 17 participants had basic technological skills (e.g., they regularly used electronic devices such as computers and smartphones), 9 suffered from hearing problems, 16 had a basic education and 14 had a superior level of education. 18 participants had been diagnosed some cognitive impairment. Most of the users were in a positive frame of mind (14 were happy to participate and 16 simply accepted it). 22 users were clearly focused.

Regarding the white-coat effect, it is worth mentioning that 91.67\% of the participants without any cognitive impairment and 61.11\% with mild or severe cognitive impairments were relaxed during the experiments. As it could be expected beforehand, cognitive impairment level increased with age \citep{Ammal2020}. 

\subsubsection{Similarity Metric Assessment}
\label{sim_study}

Each experiment was composed of five different sessions whose corresponding news items were related to economy, politics, science, society, and sports. Each session consisted of a newscast and four questions, as explained in Section \ref{system_architecture}.

In the experiments, our system was able to
separate users in most cases by {\it sim} scores that were significantly related to their level of cognitive impairment. 
Table \ref{results_average} averages the {\it sim} metric for the three groups in the study (absent, mild and severe impairment).

\begin{table}[ht!]
\centering
\caption{\label{results_average} Average $\pm$ {\sc sd} {\it sim} metric by level of impairment across all sessions.}
\begin{tabular}{lc}\hline
{\bf Level of impairment} & {\bf Avg. $\pm$ SD {\it sim} metric}\\\hline
Absent & 0.42 $\pm$ 0.17 \\
Mild & 0.29 $\pm$ 0.17 \\
Severe & 0.08 $\pm$ 0.10\\\hline
\end{tabular}
\end{table}

By session, Table \ref{results_session} shows that users with severe cognitive impairments scored significantly less, while healthy ones or those with mild levels of cognitive impairment performed significantly better. The effect of mild cognitive impairment (compared to its absence) in the comprehension skills of the participants was noticeable in all sessions except for the third. This session was particularly challenging due to its vocabulary, which included technical terms such as {\it pymes} `SMEs' (acronym for small and medium enterprises). Furthermore, in sessions 1, 2 and 4 the difference between mild and no cognitive impairment was noteworthy.

\begin{table*}[ht!]
\centering
\caption{\label{results_session} Average $\pm$ {\sc sd} {\it sim} metric at each session by level of impairment.}
\begin{tabular}{llllll}\hline
& \multicolumn{5}{c}{\bf{Session}} \\
& \multicolumn{1}{c}{\bf{1}} & \multicolumn{1}{c}{\bf{2}} & \multicolumn{1}{c}{\bf{3}} & \multicolumn{1}{c}{\bf{4}} & \multicolumn{1}{c}{\bf{5}} \\\hline
\bf{Absent} & 0.39 $\pm$ 0.27 & 0.75 $\pm$ 0.30 & 0.12 $\pm$ 0.10 & 0.48 $\pm$ 0.43 & 0.34 $\pm$ 0.41 \\
\bf{Mild} & 0.23 $\pm$ 0.21 & 0.65 $\pm$ 0.46 & 0.11 $\pm$ 0.14 & 0.19 $\pm$ 0.38 & 0.27 $\pm$ 0.38 \\
\bf{Severe} & 0.07 $\pm$ 0.07 & 0.13 $\pm$ 0.14 & 0.01 $\pm$ 0.03 & 0.00 $\pm$ 0.00 & 0.20 $\pm$ 0.45\\\hline
\end{tabular}
\end{table*}

Session 5 was particularly interesting. In it, users received several clues in user-centred questions before being presented the attention-demanding question. Thus, users with mild cognitive impairment performed better in this session than in session 3, for example. This is coherent with the fact that the reinforcement of key ideas from the news helped users to accurately answer attention-demanding questions.

Table \ref{answers_length} shows the average lengths of user responses. They tended to be shorter for higher impairment levels. In fact, for severe impairment, users tended to remain silent or answer concisely (e.g., `I don't know').

\begin{table}[ht!]
\centering
\caption{\label{answers_length} Average answer length by level of impairment, in characters.}
\begin{tabular}{lc}\hline
{\bf Level of impairment} & {\bf Avg. length}\\\hline
Absent & 54.20 \\
Mild & 37.75 \\
Severe & 30.84\\\hline
\end{tabular}
\end{table}

Table \ref{stress} shows that, for users with cognitive impairments, stress had an appalling effect on performance, and that high focus had a very positive outcome, even enhancing $sim$ results from $0.20$ to $0.36$ in relaxed users and, remarkably, from $0.03$ to $0.26$ in stressed users.

\begin{table}[ht!]
\centering
\caption{\label{stress} Average $\pm$ {\sc sd} {\it sim} metric for users with cognitive impairments by stress and focus.}
\begin{tabular}{llll}\hline
& & \multicolumn{2}{c}{\bf{Stress}}\\
& & \multicolumn{1}{c}{No} & \multicolumn{1}{c}{{Yes}} \\\hline
\multirow{2}{*}{\bf Focus} & No & 0.20 $\pm$ 0.19 & 0.03 $\pm$ 0.00 \\
& Yes & 0.36 $\pm$ 0.14 & 0.26 $\pm$ 0.16 \\\hline
\end{tabular}
\end{table}

Finally, Table \ref{skills} presents {\it sim} measurements versus technological skills and levels of education, for users with cognitive impairments. In view of these results, more educated users performed better in the experiments than those with a basic education. Apparently, technological skills only led to higher $sim$ scores for users with a basic education (note, however, the large standard deviation of the $sim$ score of skilled users with superior educational level). 

\begin{table}[ht!]
\centering
\caption{\label{skills} Average $\pm$ {\sc sd} {\it sim} metric for users with cognitive impairments by technological skills and level of education.}
\begin{tabular}{llll}\hline
 & & \multicolumn{2}{c}{\bf{Technological skills}}\\
& & \multicolumn{1}{c}{Agnostic} & \multicolumn{1}{c}{{Skilled}} \\\hline
\multirow{2}{*}{\bf Education} & Basic & 0.26 $\pm$ 0.17 & 0.29 $\pm$ 0.17 \\
& Superior & 0.34 $\pm$ 0.08 & 0.32 $\pm$ 0.35 \\\hline
\end{tabular}
\end{table}

\subsubsection{Automatic Cognitive Impairment Detection}
\label{auto_impairment}

Finally, in order to evaluate the effectiveness of the proposed system, we trained a set of selected Machine Learning algorithms \citep{Quinlan1993} for detecting cognitive impairment: 
Bayesian Network ({\sc bn}), Decision Tree ({\sc dt}), Random Forest ({\sc rf}) and linear Support Vector Machine ({\sc svm}), which have been widely used in medical applications \citep{Lu2017,Bratic2018,Ghoneim2018,Rukmawan2021,Ahmed2021}.

Table {\ref{tab:complexity}} shows the training and classification complexity of the Machine Learning algorithms we selected, for $c$ classes, $d$ features, $k$ instances of the algorithm (where applicable) and $n$ samples. {\sc{bn}} training has linear training complexity \citep{Lu2006}. {\sc{dt}} and {\sc{rf}} training have logarithmic training complexity \citep{Witten2016,Hassine2019}. {\sc{svm}} has the highest training complexity, but it is very fast in classification time if trained with a linear kernel, as in our case \citep{Vapnik2000}. We employed the algorithm implementations from Weka\footnote{Available at {\tt https://www.cs.waikato.ac.nz/ml/weka}, September 2021.} \citep{Witten2016}. 

\begin{table*}[ht!]
\caption{\label{tab:complexity} Training and testing complexity of the Machine Learning algorithms.}
\centering
\begin{tabular}{lll}\bottomrule
\multicolumn{1}{c}{\bf Classifier} & \multicolumn{1}{c}{\bf Train complexity} & \multicolumn{1}{c}{\bf Test complexity}\\\hline
BN & O($n\cdot d$) & O($c \cdot d$)\\
DT & O($n \cdot \mbox{log}(n) \cdot d$) & O(depth of the tree)\\
RF & O($n\cdot \mbox{log}(n) \cdot d \cdot k$) & O(depth of the tree$\cdot k$)\\
SVM & O($n^2$) & O($d$)
\\\toprule
\end{tabular}
\end{table*}

Firstly, we divided the sample into user classes with and without impairments. 
Table \ref{tab:features} shows all the features we considered. Then, we applied the {\tt GainRatioAttributeEval} feature selection algorithm, also from Weka, which evaluates the relevance of the attributes by measuring their gain ratio with regard to the target class. The most relevant features it selected for the classification model were, in decreasing importance, length of the response in characters, focus, {\it sim} for question 4 in session 2, age, technology skills, and {\it sim} for question 4 in session 4.

\begin{table*}[ht!]
\centering
\small
\caption{\label{tab:features} Features for the Machine Learning models training.}
\begin{tabular}{lll}
\hline
\bf Type & \bf Feature name & \bf Description \\\hline

\multirow{4}{*}{Boolean} & Focus & \begin{tabular}[c]{@{}p{7cm}@{}} True if the user was focused during the experiments, otherwise false.\end{tabular}\\

 & Stress & \begin{tabular}[c]{@{}p{7cm}@{}} True if the user was stressed during the experiments, otherwise false.\end{tabular}\\
 
 & Studies & \begin{tabular}[c]{@{}p{7cm}@{}} True if the user had a superior level of education, otherwise false.\end{tabular}\\
 
 & Technology & \begin{tabular}[c]{@{}p{7cm}@{}} True if the user had technological skills, otherwise false.\end{tabular}\\

Nominal & Age & \begin{tabular}[c]{@{}p{7cm}@{}} $\{1, 2, 3, 4\}$ if the user age was in $\{[60,70], (70,80], (80,90], (90,100]\}$, respectively.\end{tabular}\\

\multirow{6}{*}{Numerical} & NumChars & \begin{tabular}[c]{@{}p{7cm}@{}} Avg. number of characters in the user responses.\end{tabular}\\
 
 & SimS1Q4 & \begin{tabular}[c]{@{}p{7cm}@{}} {\it sim} value for question 4 in session 1.\end{tabular}\\
 
 & SimS2Q4 & \begin{tabular}[c]{@{}p{7cm}@{}} {\it sim} value for question 4 in session 2.\end{tabular}\\

 & SimS3Q4 & \begin{tabular}[c]{@{}p{7cm}@{}} {\it sim} value for question 4 in session 3.\end{tabular}\\

 & SimS4Q4 & \begin{tabular}[c]{@{}p{7cm}@{}} {\it sim} value for question 4 in session 4.\end{tabular}\\

 & SimS5Q4 & \begin{tabular}[c]{@{}p{7cm}@{}} {\it sim} value for question 4 in session 5.\end{tabular}\\

\hline
\end{tabular}
\end{table*}

Note that in spite of the results in Table \ref{skills}, the selection algorithm preferred technological skills over level of education.

Finally, Table \ref{machine_learning_results} shows the classification results for the selected algorithms with 10-fold cross validation \citep{Berrar2019} to avoid overfitting. This methodology minimises underestimation and overestimation in the results. For this purpose, the dataset is divided into 10 segments, 9 of which are used for training and the remaining one for testing. This process is repeated ten times by avoiding overlapping testing segments in different evaluations. At the end, the final performance metric is computed as the average of the intermediate tests. In addition, unlike \textsc{svm}\citep{Jatav2018} and \textsc{bn}\citep{Wood2019} algorithms, which are less prone to overfitting issues, in the cases of \textsc{dt} and \textsc{rf} we limited the folds to 3 and the maximum depth to 5. To avoid bias, the entries from the same users were grouped to prevent them from being simultaneously used for training and evaluation.

Note that {\sc dt} was finally selected due to its better performance, since it attained a detection accuracy of $86.67$\% using the most relevant features.

\begin{table*}[!htbp]
\centering
\caption{\label{machine_learning_results}F-measure, recall and response times for the selected algorithms.}
\small
\begin{tabular}{cccccc}
\toprule
\bf Classifier & \bf Class & \bf F-measure & \bf Recall & \bf Training (ms) & \bf Testing (ms)\\ \hline

\multirow{2}{*}{BN}
& Present & 64.50\% & 55.60\% & \multirow{2}{*}{0.16}&\multirow{2}{*}{$<$0.01} \\
& Absent & 62.10\% & 75.00\% \\
\cline{2-6}

\multirow{2}{*}{DT}
& Present & \bf 88.20\% & \bf 83.30\% & \multirow{2}{*}{0.15}&\multirow{2}{*}{$<$0.01} \\
& Absent & \bf 84.60\% & \bf 91.70\% \\
\cline{2-6}

\multirow{2}{*}{RF}
& Present & 74.30\% & 72.20\% & \multirow{2}{*}{0.10}&\multirow{2}{*}{$<$0.01} \\
& Absent & 64.00\% & 66.70\%\\
\cline{2-6}

\multirow{2}{*}{SVM}
& Present & 78.80\% & 72.20\% & \multirow{2}{*}{2.58}&\multirow{2}{*}{$<$0.01} \\
& Absent & 74.10\% & 83.30\% \\
\bottomrule
\end{tabular}
\end{table*}

\section{Conclusions}
\label{conclusions}

Even though there exists previous research on intelligent systems for therapeutic monitoring of cognitive impairment in elderly people, most current approaches are based on manual tests that rely on human supervision for early detection. 

In this work, to reduce caregivers' effort and the white-coat effect, we have proposed a novel conversational system for entertainment and therapeutic monitoring of elderly people. It relies on {\sc nlp} techniques for chatbot behaviour generation and user-transparent automatic assessment, by combining distracting (user-centred) with attention-demanding questions (embedded cognitive tests). Thus, our main contribution is a Machine Learning approach for user-transparent cognitive monitoring that is embedded into a user-centred entertainment solution. This approach is based on metrics that estimate the abstraction skills of the users from their answers during the automatic dialogue stages.

Experimental results with elderly people under the supervision of {\sc afaga} gerontologists indicate that our solution is satisfactory and has strong potential for user-friendly therapeutic monitoring. Preliminary analyses have obtained a detection accuracy of cognitive impairment close to 90\%.

Given these promising results, we plan to enhance the system with empathetic capabilities through user feedback. At the same time, users will benefit from real-time encouragement about their performance.

\section*{Acknowledgments}
This work was supported by Xunta de Galicia grant {\sc grc} 2018/053, Spain; and University of Vigo/CISUG for open access charge. The authors are indebted to {\sc afaga} for providing access to real patients and expert knowledge about monitoring of cognitive impairments, and for validating our results.

\section*{Declarations}

\subsection*{Data Availability}
The datasets generated during and/or analysed during the current study are available from the corresponding author on reasonable request.

\subsection*{Competing Interests}
The authors have no competing interests to declare that are relevant to the content of this article.

\subsection*{Ethics approval}
All procedures performed in studies involving human participants were in accordance with the ethical standards of the institutional and/or national research committee and with the 1964 Helsinki declaration and its later amendments or comparable ethical standards.

\subsection*{Consent to participate}
Informed consent was obtained from all individual participants in the
study.

\bibliographystyle{plainnat}
\bibliography{sn-bibliography}

\end{document}